\documentclass[conference]{IEEEtran}
\IEEEoverridecommandlockouts
\usepackage{cite}
\usepackage{amsmath,amssymb,amsfonts}
\usepackage{algorithmic}
\usepackage{graphicx}
\usepackage{textcomp}
\usepackage{xcolor}
\def\BibTeX{{\rm B\kern-.05em{\sc i\kern-.025em b}\kern-.08em
    T\kern-.1667em\lower.7ex\hbox{E}\kern-.125emX}}

\usepackage{tabularx}
\usepackage[linesnumbered,ruled]{algorithm2e}

\begin{document}

\title{Multilingual Abusiveness Identification on Code-Mixed Social Media Text}

\author{\IEEEauthorblockN{Ekagra Ranjan}
\IEEEauthorblockA{Microsoft \\
eranjan@microsoft.com}
\and
\IEEEauthorblockN{Naman Poddar}
\IEEEauthorblockA{Indian Institute Of Technology (IIT) Guwahati \\
namanpoddar@alumni.iitg.ac.in}

%\author{\IEEEauthorblockN{Anonymous}
}

\maketitle

\begin{abstract}
Social Media platforms have been seeing adoption and growth in their usage over time. This growth has been further accelerated with the lockdown in the past year when people's interaction, conversation, and expression were limited physically. It is becoming increasingly important to keep the platform safe from abusive content for better user experience. Much work has been done on English social media content but text analysis on non-English social media is relatively underexplored. Non-English social media content have the additional challenges of code-mixing, transliteration and using different scripture in same sentence. In this work, we propose an approach for abusiveness identification on the multilingual Moj dataset which comprises of Indic languages. Our approach tackles the common challenges of non-English social media content and can be extended to other languages as well. 
\end{abstract}

\section{Introduction}
Social media platorms have been seeing growing usage and adoption over time. This growth has been further accelerated with the lockdowns in the pandemic where people had no choice but to interact and express themselves online. This has led to increase in new users as well as daily active users. India has seen tremendous adoption in social media usage with the Jio revolution when Reliance Jio made internet charges affordable. Reliance Jio allowed democratizing internet which has benefitted social media platforms operating in India. With the rise of intenet access to Indian people, social media platforms are seeing rise in Indic native content. It is in interest of social media platforms to adapt their systems to understand the regional content to better serve the users and attract more user base. One such company which has been focusing on serving the regional content since its foundation is Sharechat. Sharechat is an Indian company which is one of the leading social networking space in catering to regional content. Their short video app is called Moj which went viral in the lockdown and has been since seeing tremendous growth in its user base and content. 

Increasing social media usage has made it important to keep the platform safe from abusive content to enable individual safety in the online world. There is more abusive content in the online world when compared to the offline world because Internet enables anonymity. This can be used to share abusive content without being held responsible. Text analysis of English social media content is a well explored domain \cite{sentiment-svm}, \cite{sentiment}. Multilingual NLP is a relative new subfield and has been picking up pace due to globalization \cite{xlm-r-twitter}. Most of the multilingual transfomer models were trained on Wikipedia dataset of Common Crawl dataset \cite{mbert}, \cite{xlm-r}. This is different from  training on a social media text dataset as social media text content have vast amount of acceptable misspellings. On top of this, non-English social media content have their own unique set of challenges comprising of code-mixing, transliteration, mixing of scriptures in the same sentences. Code-Mixing is the usage of different languages in the same sentence or utterance. This mostly used in informal conversation and social media. This phenonmenon exits in multilingual societies where the speaker is fluent in more than one language \cite{bertologicomix}. Transliteration is a way of writing a native language in the scripture of a different language. The pronunciation of the transliterated langugae, by following the rules of the scripture used, is similar to the one written using the native scripture. Most common form of transliteration is to use the Latin scripture to express a non-English sentence. Transliteration aggravates the misspellings and variants of the same word. This is because not all syllables and sounds in the native language would be available in the Latin language so the writer has to use non-standard ways to approximate the syllable. 

Prior works have performed text analysis on non-English languages and have attempted to tackle a subset of the above challenges associated with non-English social media text \cite{2020-semeval}. \cite{sentiment-arabic}, \cite{sentiment-czech}, and \cite{sentiment-telegu} have extended text analysis to Arabic, Czech, and Telugu languages. In our work, we identify the common challenges associated with non-English social media text and propose an approach to tackle all of these challenges in multilingual setting. Our contributions include a spell-correction algorithm that does not rely on a manually defined dictionary but instead infers its own from the training data. This allows our spell-correction algorithm to learn new words by retraining on the any new corpus. This is in contrast to the traditional dictionary based approach where the words need to be added manually. Our approach is scalable in the age of neologism which is prevalant in context of social media \cite{neologism}.

% section overview
The paper is organized as follows. Sec. \ref{sec:dataset} describes the Moj Dataset used in this work. Sec \ref{sec:problem-statement} defines the problem statement and challenges associated with it. Sec. \ref{sec:proposed-method} explains our approach in detail. Sec. \ref{sec:exp-setup} provides the experimental setup. Sec. \ref{sec:result} lists the result of our approach and in  \ref{sec:ablation} we perform ablation studies on our approach. Sec. \ref{sec:future-work} discusses the future work and we conclude our work in Sec. \ref{sec:conclusion}

\section{Dataset}
\label{sec:dataset}
The dataset provided by Moj consists of multilingual comments in $10+$ low-resource Indic languages. The comments in the dataset have been human annotated with a label indicating whether the comment is abusive or not. The dataset also has \textit{meta-data} like the \textit{language} of the comment and the number of \textit{likes} and \textit{reports} on each comment and the corresponding post on which the comment was made. The label distribution across different language is show in Table \ref{tab:label-dist} 

\begin{table}[h]
	\centering
	\caption{Label distribution across languages in Moj Dataset}
	\label{tab:label-dist}
	
	\bgroup
	\def\arraystretch{1.2}%  1 is the default, change whatever you need
	\begin{tabular}{ccc}
		\hline
		\textbf{Language} & \textbf{Non-Abusive} & \textbf{Abusive}\\
		\hline
		Assamese & 1496 & 1284\\
		Bengali & 11428 & 11407\\
		Bhojpuri & 2917 & 2887\\
		Gujarati & 4426 & 4402\\
		Haryanvi & 4395 & 4417\\
		Hindi & 153433 & 153747\\
		Kannada & 6954 & 6989\\
		Malayalam & 31749 & 9216\\
		Marathi & 44677 & 27367\\
		Odia & 5475 & 5499\\
		Rajasthani & 2183 & 2185\\
		Tamil & 34792 & 34705\\
		Telugu & 48461 & 48551\\
		
		\hline
	\end{tabular}
	\egroup
\end{table}

\section{Problem Statement}
\label{sec:problem-statement}
\begin{figure}[!ht]
	\includegraphics[width=0.43\textwidth]{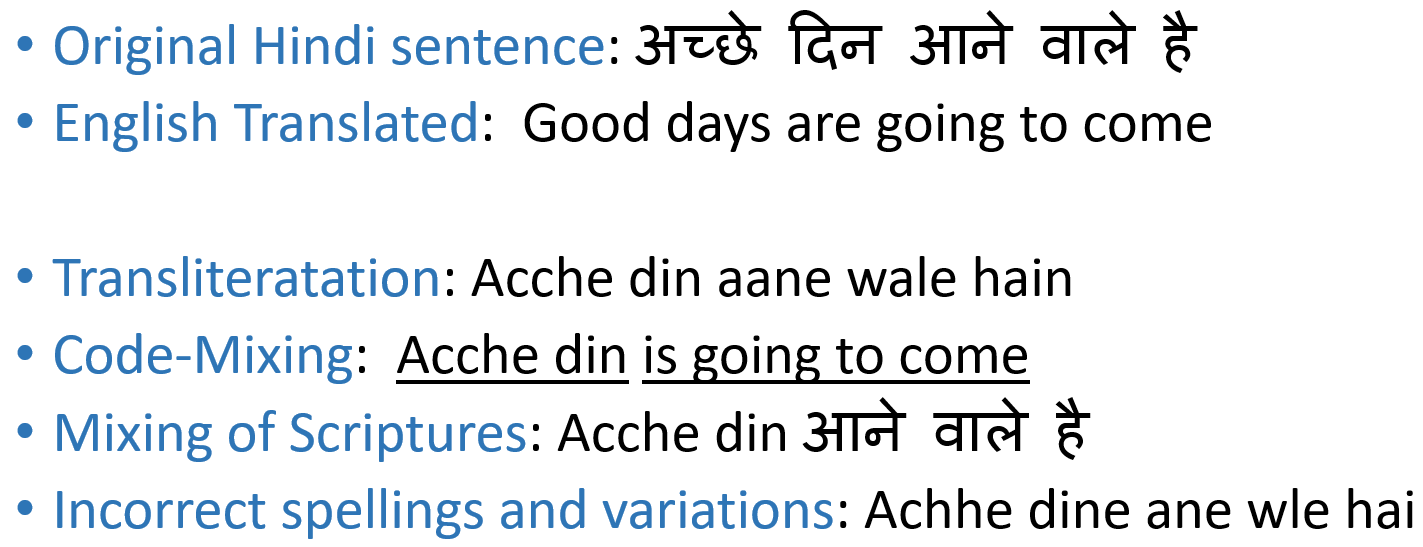}
	\caption{\label{fig:challenge} Example of challenges described in Sec. \ref{sec:problem-statement}.}
\end{figure}

The main objective of the paper is to develop a multi-lingual approach to identify abusive comments for social media platforms. Upon exploring the Moj dataset, we identify the following challenges associated with multi-lingual non-English social media content:
\begin{itemize}
	\item \textbf{Code-Mixing}: Social media text is not written in just one language in multi-lingual socities.
	\item \textbf{Transliteration}: People often use Latin scripture to write sentences in their native language. Pronunciation of both original native sentence and transliterated sentence is almost similar.
	\item \textbf{Mixing of Scriptures}: Some people write sentences which contain words from different scriptures.
	\item \textbf{Misspelling and variations} in social media text.
\end{itemize}
Fig. \ref{fig:challenge} shows different representation of the same Hindi sentence. Our approach attempts to tackle such challenges in multi-lingual setting.

\section{Proposed Method}
\label{sec:proposed-method}
\subsection{Data Cleaning}
\label{ssec:data-cleaning}
The dataset includes multiple comments from the same post as well. We replace the post level meta-data of each comment, i.e., post like count and post report count, with the maximum value of respective meta-data feature among all the comments corresponding to the same post.

We pre-process the comments in the Moj dataset with the following steps:
\begin{itemize}
	\item Unicode normalization using \verb|unicodedata| python package
	\item Remove special characters, e.g, $@$, $"$, $\$$, $\#$
	\item Remove Emoji using \verb|demoji|\footnote{https://pypi.org/project/demoji/} python package
	\item Replace characters which occur more than twice consecutively with the same character, e.g., hellooo \textrightarrow hello
	\item Transliterate the comments using \verb|indic-trans| python package \cite{indic-trans} to the native laguage scripture to tackle the challenge of transliteration mentioned in Sec. \ref{sec:problem-statement}. The effect of this pre-processing step is discussed in Sec. \ref{sec:ablation}.
\end{itemize}

\begin{figure*}[!ht]
	\centering
	\includegraphics[width=1\textwidth]{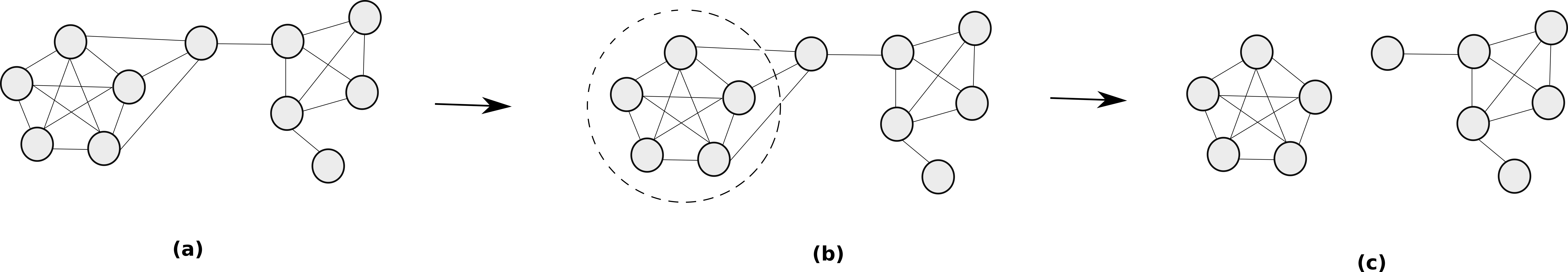}
	\caption{\label{fig:spell-correct} Overview of the training phase of Spell Correction using Graph Clustering: 
		(a) Graph is constructed using words in the training corpus. Two words have an edge in the graph if they satisfy the nearness criteria specified in Sec. \ref{sssec:train-spell-correct}.
		(b) Maximal clique is found in this graph. The circled subgraph in the figure is the maximal clique in this graph. This subgraph is considered to contain words that are misspellings of one another and are representing the same concept.
		(c) This subgraph/cluster is removed from the graph and is processed to obtain its \textit{Parent Word} and \textit{Anchor Words} as explained in Sec. \ref{sssec:train-spell-correct}. The whole process is again repeated using the remaining smaller graph untill all the nodes have been processed. Please refer to Sec.
		\ref{ssec:spell-correct} for more details.}
\end{figure*}

\subsection{Spell Correction using Graph Clustering}
\label{ssec:spell-correct}
Our approach builds a graph of words using the training corpus of Moj dataset. Sec. \ref{sssec:train-spell-correct} shows how graph clustering is used to identify the set of related words and the correct spelling of those words. Sec. \ref{sssec:test-spell-correct} shows the strategy used to identify the correct spelling. This approach works even for the wrong words that were not seen in the training as long as the correct word was seen in training. Sec. \ref{sec:ablation} shows the utility of our approach.
\subsubsection{Training: Correct Spelling Identification}
\label{sssec:train-spell-correct}
We intend to spell-correct the abusive words and ignore words which occur predominantly in non-abusive comments. This is done to prevent non-abusive words from being auto-corrected to abusive words or vice-versa. Since this is a classification problem, presence of abusive word is more important than presence of non-abusive word. This is because any combination of abusive and non-abusive word will be abusive in nature. 

Consider words that have occured minimum $5$ times in the abusive comments in the training corpus. Ignore words that have length less than $6$ and have number of consonants less than $4$.  Construct a graph where nodes are these words. Connect words with an edge if they have the same first $k$ letters and the Levenshtein Distance between them is above a threshold score $t$. In this work, $k$ was $3$ and score $t$ was $85$ based on hyper-parameter tuning on a validation set. Levenshtein Distance was computed using \verb|rapidfuzz|\footnote{https://pypi.org/project/rapidfuzz/}. Algorithm \ref{alg:Graph} is a psuedo-code of the graph construction.

Find the maximal clique from the above constructed graph of words. A maximal clique is the largest subgraph where nodes are connected to one another by an edge. This maximal clique represents a subgraph of words which our algorithm thinks are variants or misspellings of one another and represent the same thing, e.g, \verb|bewakoof|, \verb|bewakoofi|, \verb|bewakouf| where \verb|bewakoof| means idiot in Hindi. The word in this cluster with highest occurring frequency in the training corpus is chosen as the correct spelling for this cluster. We refer to this word as the \textit{Parent Word} of the cluster. Next, we find the \textit{Anchor Words} for this cluster. \textit{Anchor Words} are those words which will represent this cluster during Testing phase explained in Sec. \ref{sssec:test-spell-correct}. Top five frequent words of this cluster are chosen as the candidates for the \textit{Anchor Words}. Among them, those words that have frequency above one-fourth of the the \textit{Parent Word} is included in \textit{Anchor Words}.

The above maximal clique is removed from the graph once that cluster is processed and its \textit{Parent Words} and \textit{Anchor Words} have been computed. This process of finding a cluster using maximal clique, processing the cluster, and removing it is repeated on the remaining smaller graph obtained from the previous rounds of clustering until all the nodes in the original graph have been processed. Algorithm \ref{alg:train} shows the psuedo-code for training phase.

\subsubsection{Testing: Auto-correction during inference}
\label{sssec:test-spell-correct}
By the end of training phase, we have a set of clusters with \textit{Parent Word} and its \textit{Anchor Words} that will be used during inference to spell-correct the words in train and test corpus. Given a word $w$ during inference, we find out the clusters whose \textit{Parent Word}  has the same first $k$ letters as that of word $w$. The association $score_{wi}$ of a cluster $C_{i}$ obtained from the previous filtering step is given as:
\begin{equation}
\label{eq:y}
score_{wi} = \max_{a \in A(C_{i})} \{Levenshtein Distance(w, a)\}
\end{equation}
where $A(C_{i})$ is the set of \textit{Anchor Words} of cluster $C_{i}$. The cluster $C_{j}$ that has the highest association $score_{wj}$ with the word $w$ is chosen as the appropriate cluster $C^{*}$ if the score $score_{wj}$ is above a threshold $t$. The word $w$ is replaced with the \textit{Parent Word} of the cluster $C^{*}$ if the word $w$ was able to find its appropriate cluster $C^{*}$, e.g., \verb|bewakoufi| will be corrected to \verb|bewakoof|. We spell-correct all the comments in both training and testing data before using them for classification. Algorithm \ref{alg:test} shows the psuedo-code for the testing phase.

\subsection{Multilingual Models}
\label{ssec:model}
Given that our task involves low-resource Indic languages, we use multiple pre-trained multilingual transformer models and finetune them on the comments of Moj dataset. The following models were used in our work:
\begin{itemize}
	\item \textbf{mBERT} \cite{mbert}: is a multilingual variant of BERT \cite{bert} trained on multilingual Wikipedia data from $104$ languages.
	\item \textbf{XLM-R} \cite{xlm-r}: is trained on cross-lingual masked language modeling task on $100$ languages.
	\item \textbf{XLM-T} \cite{xlm-r-twitter}: is XLM-R trained on Twitter comments
	\item \textbf{Multilingual DistilBERT} \cite{distilbert}: is a distilled version of mBERT
	\item \textbf{MURIL} \cite{muril}: is a variant of BERT trained on monolingual, translated and transliterated data from $17$ Indic languages
	\item \textbf{Indic-BERT} \cite{indic-bert}: is a variant of ALBERT \cite{albert} trained on IndicCorp dataset \cite{indic-bert}.
	\item \textbf{Canine} \cite{canine}: is a tokenizer and vocabulary free transformer model which operates directly on sequences of characters. Canine-S and Canine-C are two variants trained on different loss functions defined in \cite{canine}.
\end{itemize}

We also trained a logistic regression model on an embedding created by using Naive-Bayes over TF-IDF feautures \cite{nb-svm}\footnote{https://www.kaggle.com/jhoward/nb-svm-strong-linear-baseline}.

\subsection{Meta-Data}
The meta-data associated with the comments, i.e, likes and report count of the comments as well as the post on which the comment was made, was leveraged to further improve the predictions. We trained a Random Forest classifier over this meta-data only to provide the predictions. These meta-data specific predictions are later combined with the predictions of the comment-specific models as discussed later in Sec. \ref{sec:result}.

%algo
\begin{algorithm}[!ht]
	\caption{Graph Construction}
	\label{alg:Graph}
	\SetKwInOut{Input}{Input}\SetKwInOut{Output}{Output}
	\Input{~Words $W$ present in abusive sentences; Frequency count $F$ containing number of times a word has occurred in the corpus; Length of words $L$; Number of consonants in words $C$; Function $T(w)$ gives the first three letters of input word $w$; Levenshtein Distance function $score(a, b)$}
	\Output{~Word Graph $G(V, E)$ where $V$ is the nodes and $E$ is the edges in the graph $G$}
	\BlankLine
	$V \leftarrow \phi$ \\
	\For{$i=1...|W|$}
	{
		\If{$F_{i}\geq 5$}
		{
			\If{$L_{i} \geq 6$}
			{
				\If{$C_{i} \geq 4$}
				{
					\If{$V = \phi$}
					{
						$V_{1} \leftarrow W_{i}$
					}
					\Else
					{
						\For{$j=1...|V|$}
						{
							\If{$T(W_{i}) = T(V_{j})$}
							{
								\If{$score(W_{i}, V_{j}) \ge 85$}
								{
									$k \leftarrow |V| + 1$\\
									$V_{k} \leftarrow W_{i}$\\
									
									$E_{j} \leftarrow E_{j} + V_{k}$\\
									$E_{k} \leftarrow E_{k} + V_{j}$\\
								}
							}
						}
					}
				}
			}
		}
	}
\end{algorithm}

%\onecolumn
\begin{algorithm}[!ht]
	
	\caption{\textsc{Training: Correct Spelling Identification} algorithm}
	\label{alg:train}
	\SetKwInOut{Input}{Input}\SetKwInOut{Output}{Output}\SetKwInOut{Intermediate}{Intermediate}
	\Input{~Graph $G(V,E)$; Frequency count $F$ containing number of times a word has occurred in the corpus}
	\Intermediate{~$C$ is the maximal cluster obtained in a particular step; $p$ is the parent element for a particular cluster; $a$ is the anchor elements for a particular cluster}
	\Output{~Parent $P$ containing list of parent member for the clusters that will be obtained; Anchor $A$ containing list of anchor members for the clusters that will be obtained}
	\BlankLine
	$G' \leftarrow G$\\
	$A \leftarrow \phi$\\
	$P \leftarrow \phi$\\
	
	\While{$G' \ne \phi$}
	{
		$C(V^{c}, E^{c}) = MaximalClique(G')$\\
		$j = argmax(\{F_{i} | i \in \{1...|V^{c}|\} \})$\\
		$p \leftarrow V^{c}_{j}$\\
		$p_{freq} = F_{j}$\\
		$a \leftarrow \phi$\\
		\For{$k \in \{1...|V^{c}|\}\}$}
		{
			\If{$F_{k} > \frac{p_{freq}}{4}$}
			{
				$a \leftarrow a + V^{c}_{k}$\\
			}
		$G' \leftarrow G' - C$\\
		$P \leftarrow P + p$\\
		$A \leftarrow A + a$\\
		}
	}
	
\end{algorithm}

\begin{algorithm}[!ht]
	
	\caption{\textsc{Testing: Auto-Correction duing Inference} algorithm}
	\label{alg:test}
	\SetKwInOut{Input}{Input}\SetKwInOut{Output}{Output}
	\Input{~$w_{T}$ is the test word that needs to be corrected; Parent $P$ containing list of parent member for the clusters obtained during training; Anchor $A$ containing list of anchor members for the clusters obtained during training; $score(a, b)$ is the Levenshtein distance between words $a$ and $c$; Function $T(w)$ gives the first three letters of input word $w$}
	\Output{~$w^{*}$ is the corrected word for $w_{T}$}
	\BlankLine
	$w^{*} = w_{T}$\\
	$score^{*} = 0$\\
	\For{$i \in \{1...|P|\}$}
	{
		\If{$T(w_{T}) = T(P_{i})$}
		{
			\For{$j \in {1...|A_{i}|}$}
			{
				$s = score(w_{T}, A_{i,j})$\\
				\If{$s \ge 85$}
				{	
					\If{$s \ge score^{*}$}
					{
						$s = score^{*}$\\
						$w^{*} = P_{i}$ \\
					}
				}
			}
		}
	}
	
\end{algorithm}

\section{Experimental Setup}
\label{sec:exp-setup}
All transformer models except for Canine are finetuned using Adam optimizer \cite{adam} with learning rate of $1e$-$5$. Canine models were finetuned using AdamW optimizer \cite{adamw} using $2e$-$5$ learning rate. All the models were trained for $3$ epochs with binary-cross entropy as the loss function. Threshold of $0.5$ was used to get the output class from output probabilites. Huggingface library \cite{huggingface} was used for training all the transformer models.

\section{Result}
\label{sec:result}

\begin{table}[h]
	\centering
	\caption{Comparison of Models on Moj Dataset}
	\label{tab:model-result}
	
	\bgroup
	\def\arraystretch{1.2}%  1 is the default, change whatever you need
	\begin{tabular}{ccc}
		\hline
		\textbf{Model} & \textbf{Val. F1} & \textbf{Test F1}\\
		\hline
		MURIL & \textbf{0.8675} & - \\
		Indic-BERT & 0.8475 & - \\
		XLM-R & 0.8623 & - \\
		XLM-T & 0.8633 & - \\
		Multilingual DistilBERT & 0.8553 & - \\
		mBERT & 0.8551 & - \\
		Canine-C & 0.8646 & - \\
		Canine-S & 0.8622 & - \\
		Logistic Regression + NB + TF-IDF & 0.8559  & - \\
		Random Forest (meta-data) & 0.7109 & - \\
		Ensemble & - & \textbf{0.8933}\\ 
		
		\hline
	\end{tabular}
	\egroup
\end{table}

We create a stratified validation set from $10\%$ of the training set. Table \ref{tab:model-result} shows the comparison of all the models discussed in Sec. \ref{ssec:model}. Logistic Regression + Naive Bayed on TF-IDF provides a good baseline even though it is a very simple model. MURIL scores the highest F1 score on the validation set. XLM-T performed slightly better than XLM-R as XLM-T is trained on multilingual Twitter comment which is closer to the Moj Dataset than the Wikipedia dataset on which XLM-R was trained on. The meta-data contains signals essential for this task as the Random Forest obtains an F1 score of $0.71$ using meta-data alone.  Our final model is an ensemble of all the models from all the $3$ epochs in Sec. \ref{ssec:model}. A logistic regression model was trained on the validation set to combine the predictions from all models for ensembling.

\section{Ablation Analysis}
\label{sec:ablation}

\begin{table}[h]
	\centering
	\caption{Effect of Data Cleaning, Native Transliteration, and Spell correction using Logistic Regression}
	\label{tab:ablation}
	
	\bgroup
	\def\arraystretch{1.2}%  1 is the default, change whatever you need
	\begin{tabular}{cc}
		\hline
		\textbf{Model} & \textbf{Val. F1}\\
		\hline
		Raw Dataset & 0.8493 \\
		Cleaned Dataset & 0.8508 \\
		Cleaned + Native Transliteration Dataset & 0.8547 \\
		Cleaned + Native Transliteration + Spell-Corrected Dataset & \textbf{0.8567} \\
		Cleaned + English Transliteration Dataset & 0.8496 \\ 
		\hline
	\end{tabular}
	\egroup
\end{table}

\subsection{Data Cleaning}
Logistic regression was used to measure the effect of our pre-processing steps as its simple to train, test and iterate. Table. \ref{tab:ablation} measures the validation F1 score on raw data and the various pre-processing steps. It shows that using the data cleaning as specified in Sec. \ref{ssec:data-cleaning} improves the F1 score. 

\subsection{Transliteration}
\cite{how-multilingual-is-mbert} shows that mBERT is unable to perform good on transliterated dataset as it was not trained on such data. Since only MURIL was originally trained on transliterated data, other transformer models would face domain mismatch when finetuned on Moj Dataset. To tackle transliteration and mixture of scriptures in same sentences as discussed on Sec. \ref{sec:problem-statement}, we experimented with transliterating the comments before feeding it to our classifiers. Table. \ref{tab:ablation} shows that transliterating all the comments to the native scripture shows the largest improvement among all the steps. We also experimented with transliterating everything to English so that there is knowledge sharing between languages that have common words but have different scriptures. This would increase the number of training data for lower resource language. However, this led to a decrease in F1 score. Transliterating eveything to native might be working better than transliterating to English because there could be same sounding words that could mean different things in different language, e.g, \verb|kutta| means "dog" in Hindi whereas it means "new" in Telugu. Transliterating to English also leads to loss of syllables which are not present in English language but are present in the native language.
% chata umbrella slap

\subsection{Spell Correction}
Spelling correction allows the number of training samples per concept to increase as different variants of words are mapped to the same underlying concept, e.g, \verb|bakwas| and \verb|bakvas| where \verb|bakwas| means "useless" in Hindi. Increased number of training samples per concept allows the model to learn better representation for that particular concept. Table. \ref{tab:ablation} indicates that our clustering approach in Sec. \ref{ssec:spell-correct} boosts the F1 score.
% golden dataset

\section{Future Work}
\label{sec:future-work}
Table. \ref{tab:label-dist} shows that there is class imbalance for Malayalam language. Dravidian-CodeMix-FIRE \cite{dravidiancodemix2021-overview} can be used to augment Moj Dataset to overcome this. Adversarial training \cite{miyato2016adversarial}, \cite{pk-nutcracker}, \cite{pk-paw} can be used to increase robustness of the system. Emojis could be replaced with words instead of completely removing them as done in Sec. \ref{ssec:data-cleaning}. Incorporating homophones in the graph construction in Sec. \ref{ssec:spell-correct} might bring more gains. 

\section{Conclusion}
\label{sec:conclusion}
In this paper, we propose an approach to identify abusive comments from a multilingual dataset. We leverage pre-trained multilingual transformer models along with classical machine learning models for abusive identification. We identify the inherent challenges associated with this task such as code-mixing, transliteration, mixing of scriptures, and misspellings. Our approach attempts to tackle all of these challenges to build a robust system. We propose a spell correction algorithm using graph clustering that  does not need manual creation of a dictionary for each language. It can also correct unseen incorrect words during inference if the correct word was observed in training. Results shows that our approach is well-suited for non-English social media text analysis.

\bibliographystyle{./IEEEtran}
\bibliography{./IEEEabrv,./IEEEexample.bib}

\begin{thebibliography}{10}
\providecommand{\url}[1]{#1}
\csname url@samestyle\endcsname
\providecommand{\newblock}{\relax}
\providecommand{\bibinfo}[2]{#2}
\providecommand{\BIBentrySTDinterwordspacing}{\spaceskip=0pt\relax}
\providecommand{\BIBentryALTinterwordstretchfactor}{4}
\providecommand{\BIBentryALTinterwordspacing}{\spaceskip=\fontdimen2\font plus
\BIBentryALTinterwordstretchfactor\fontdimen3\font minus
  \fontdimen4\font\relax}
\providecommand{\BIBforeignlanguage}[2]{{%
\expandafter\ifx\csname l@#1\endcsname\relax
\typeout{** WARNING: IEEEtran.bst: No hyphenation pattern has been}%
\typeout{** loaded for the language `#1'. Using the pattern for}%
\typeout{** the default language instead.}%
\else
\language=\csname l@#1\endcsname
\fi
#2}}
\providecommand{\BIBdecl}{\relax}
\BIBdecl

\bibitem{sentiment-svm}
\BIBentryALTinterwordspacing
T.~Mullen and N.~Collier, ``Sentiment analysis using support vector machines
  with diverse information sources,'' in \emph{Proceedings of the 2004
  Conference on Empirical Methods in Natural Language Processing}.\hskip 1em
  plus 0.5em minus 0.4em\relax Barcelona, Spain: Association for Computational
  Linguistics, Jul. 2004, pp. 412--418. [Online]. Available:
  \url{https://aclanthology.org/W04-3253}
\BIBentrySTDinterwordspacing

\bibitem{sentiment}
A.~Samuels and J.~Mcgonical, ``Sentiment analysis on social media content,''
  \emph{arXiv preprint arXiv:2007.02144}, 2020.

\bibitem{xlm-r-twitter}
F.~Barbieri, L.~E. Anke, and J.~Camacho-Collados, ``Xlm-t: A multilingual
  language model toolkit for twitter,'' \emph{arXiv preprint arXiv:2104.12250},
  2021.

\bibitem{mbert}
\BIBentryALTinterwordspacing
J.~Devlin, ``Multilingual bert readme,'' 2018. [Online]. Available:
  \url{https://github.com/google-research/bert/commits/master/multilingual.md}
\BIBentrySTDinterwordspacing

\bibitem{xlm-r}
A.~Conneau, K.~Khandelwal, N.~Goyal, V.~Chaudhary, G.~Wenzek, F.~Guzm{\'a}n,
  E.~Grave, M.~Ott, L.~Zettlemoyer, and V.~Stoyanov, ``Unsupervised
  cross-lingual representation learning at scale,'' in \emph{Proceedings of the
  58th Annual Meeting of the Association for Computational Linguistics}.\hskip
  1em plus 0.5em minus 0.4em\relax Association for Computational Linguistics,
  Jul. 2020.

\bibitem{bertologicomix}
S.~Santy, A.~Srinivasan, and M.~Choudhury, ``{BERT}ologi{C}o{M}ix: How does
  code-mixing interact with multilingual {BERT}?'' in \emph{Proceedings of the
  Second Workshop on Domain Adaptation for NLP}.\hskip 1em plus 0.5em minus
  0.4em\relax Kyiv, Ukraine: Association for Computational Linguistics, Apr.
  2021.

\bibitem{2020-semeval}
\BIBentryALTinterwordspacing
P.~Patwa, G.~Aguilar, S.~Kar, S.~Pandey, S.~PYKL, B.~Gamb{\"a}ck,
  T.~Chakraborty, T.~Solorio, and A.~Das, ``{S}em{E}val-2020 task 9: Overview
  of sentiment analysis of code-mixed tweets,'' in \emph{Proceedings of the
  Fourteenth Workshop on Semantic Evaluation}.\hskip 1em plus 0.5em minus
  0.4em\relax Barcelona (online): International Committee for Computational
  Linguistics, Dec. 2020, pp. 774--790. [Online]. Available:
  \url{https://aclanthology.org/2020.semeval-1.100}
\BIBentrySTDinterwordspacing

\bibitem{sentiment-arabic}
H.~S. Ibrahim, S.~M. Abdou, and M.~Gheith, ``Sentiment analysis for modern
  standard arabic and colloquial,'' \emph{arXiv preprint arXiv:1505.03105},
  2015.

\bibitem{sentiment-czech}
\BIBentryALTinterwordspacing
I.~Habernal, T.~Pt{\'a}{\v{c}}ek, and J.~Steinberger, ``Sentiment analysis in
  {C}zech social media using supervised machine learning,'' in
  \emph{Proceedings of the 4th Workshop on Computational Approaches to
  Subjectivity, Sentiment and Social Media Analysis}.\hskip 1em plus 0.5em
  minus 0.4em\relax Atlanta, Georgia: Association for Computational
  Linguistics, Jun. 2013, pp. 65--74. [Online]. Available:
  \url{https://aclanthology.org/W13-1609}
\BIBentrySTDinterwordspacing

\bibitem{sentiment-telegu}
\BIBentryALTinterwordspacing
S.~S. Mukku, N.~Choudhary, and R.~Mamidi, ``Enhanced sentiment classification
  of telugu text using ml techniques,'' in \emph{SAAIP@IJCAI}, 2016, pp.
  29--34. [Online]. Available: \url{http://ceur-ws.org/Vol-1619/paper5.pdf}
\BIBentrySTDinterwordspacing

\bibitem{neologism}
\BIBentryALTinterwordspacing
Q.~W{\"u}rschinger, M.~F. Elahi, D.~Zhekova, and H.-J. Schmid, ``Using the web
  and social media as corpora for monitoring the spread of neologisms. the case
  of {`}rapefugee{'}, {`}rapeugee{'}, and {`}rapugee{'}.'' in \emph{Proceedings
  of the 10th Web as Corpus Workshop}.\hskip 1em plus 0.5em minus 0.4em\relax
  Berlin: Association for Computational Linguistics, Aug. 2016, pp. 35--43.
  [Online]. Available: \url{https://aclanthology.org/W16-2605}
\BIBentrySTDinterwordspacing

\bibitem{indic-trans}
\BIBentryALTinterwordspacing
I.~A. Bhat, V.~Mujadia, A.~Tammewar, R.~A. Bhat, and M.~Shrivastava, ``Iiit-h
  system submission for fire2014 shared task on transliterated search,'' in
  \emph{Proceedings of the Forum for Information Retrieval Evaluation}, ser.
  FIRE '14.\hskip 1em plus 0.5em minus 0.4em\relax New York, NY, USA: ACM,
  2015, pp. 48--53. [Online]. Available:
  \url{http://doi.acm.org/10.1145/2824864.2824872}
\BIBentrySTDinterwordspacing

\bibitem{bert}
J.~Devlin, M.-W. Chang, K.~Lee, and K.~Toutanova, ``{BERT}: Pre-training of
  deep bidirectional transformers for language understanding,'' in
  \emph{Proceedings of the 2019 Conference of the North {A}merican Chapter of
  the Association for Computational Linguistics: Human Language Technologies,
  Volume 1 (Long and Short Papers)}.\hskip 1em plus 0.5em minus 0.4em\relax
  Minneapolis, Minnesota: Association for Computational Linguistics, Jun. 2019.

\bibitem{distilbert}
V.~Sanh, L.~Debut, J.~Chaumond, and T.~Wolf, ``Distilbert, a distilled version
  of bert: smaller, faster, cheaper and lighter,'' \emph{The 5th Workshop on
  Energy Efficient Machine Learning and Cognitive Computing}, 2019.

\bibitem{muril}
S.~Khanuja, D.~Bansal, S.~Mehtani, S.~Khosla, A.~Dey, B.~Gopalan, D.~K. Margam,
  P.~Aggarwal, R.~T. Nagipogu, S.~Dave \emph{et~al.}, ``Muril: Multilingual
  representations for indian languages,'' \emph{arXiv preprint
  arXiv:2103.10730}, 2021.

\bibitem{indic-bert}
D.~Kakwani, A.~Kunchukuttan, S.~Golla, N.~Gokul, A.~Bhattacharyya, M.~M.
  Khapra, and P.~Kumar, ``inlpsuite: Monolingual corpora, evaluation benchmarks
  and pre-trained multilingual language models for indian languages,'' in
  \emph{Proceedings of the 2020 Conference on Empirical Methods in Natural
  Language Processing: Findings}, 2020, pp. 4948--4961.

\bibitem{albert}
\BIBentryALTinterwordspacing
Z.~Lan, M.~Chen, S.~Goodman, K.~Gimpel, P.~Sharma, and R.~Soricut, ``Albert: A
  lite bert for self-supervised learning of language representations,'' in
  \emph{International Conference on Learning Representations}, 2020. [Online].
  Available: \url{https://openreview.net/forum?id=H1eA7AEtvS}
\BIBentrySTDinterwordspacing

\bibitem{canine}
J.~H. Clark, D.~Garrette, I.~Turc, and J.~Wieting, ``Canine: Pre-training an
  efficient tokenization-free encoder for language representation,''
  \emph{arXiv preprint arXiv:2103.06874}, 2021.

\bibitem{nb-svm}
\BIBentryALTinterwordspacing
S.~Wang and C.~Manning, ``Baselines and bigrams: Simple, good sentiment and
  topic classification,'' in \emph{Proceedings of the 50th Annual Meeting of
  the Association for Computational Linguistics (Volume 2: Short
  Papers)}.\hskip 1em plus 0.5em minus 0.4em\relax Jeju Island, Korea:
  Association for Computational Linguistics, Jul. 2012, pp. 90--94. [Online].
  Available: \url{https://aclanthology.org/P12-2018}
\BIBentrySTDinterwordspacing

\bibitem{adam}
D.~Kingma and J.~Ba, ``Adam: A method for stochastic optimization,''
  \emph{International Conference on Learning Representations}, 12 2014.

\bibitem{adamw}
\BIBentryALTinterwordspacing
I.~Loshchilov and F.~Hutter, ``Decoupled weight decay regularization,'' in
  \emph{7th International Conference on Learning Representations, {ICLR} 2019,
  New Orleans, LA, USA, May 6-9, 2019}.\hskip 1em plus 0.5em minus 0.4em\relax
  OpenReview.net, 2019. [Online]. Available:
  \url{https://openreview.net/forum?id=Bkg6RiCqY7}
\BIBentrySTDinterwordspacing

\bibitem{huggingface}
T.~Wolf, L.~Debut, V.~Sanh, J.~Chaumond, C.~Delangue, A.~Moi, P.~Cistac,
  T.~Rault, R.~Louf, M.~Funtowicz \emph{et~al.}, ``Huggingface's transformers:
  State-of-the-art natural language processing,'' \emph{arXiv preprint
  arXiv:1910.03771}, 2019.

\bibitem{how-multilingual-is-mbert}
T.~Pires, E.~Schlinger, and D.~Garrette, ``How multilingual is multilingual
  {BERT}?'' in \emph{Proceedings of the 57th Annual Meeting of the Association
  for Computational Linguistics}.\hskip 1em plus 0.5em minus 0.4em\relax
  Florence, Italy: Association for Computational Linguistics, Jul. 2019.

\bibitem{dravidiancodemix2021-overview}
R.~Priyadharshini, B.~R. Chakravarthi, S.~Thavareesan, D.~Chinnappa,
  T.~Durairaj, and E.~Sherly, ``Overview of the dravidiancodemix 2021 shared
  task on sentiment detection in tamil, malayalam, and kannada,'' in
  \emph{Forum for Information Retrieval Evaluation}, ser. FIRE 2021.\hskip 1em
  plus 0.5em minus 0.4em\relax Association for Computing Machinery, 2021.

\bibitem{miyato2016adversarial}
T.~Miyato, A.~M. Dai, and I.~Goodfellow, ``Adversarial training methods for
  semi-supervised text classification,'' \emph{arXiv preprint
  arXiv:1605.07725}, 2016.

\bibitem{pk-nutcracker}
\BIBentryALTinterwordspacing
P.~Kumar and A.~Singh, ``{N}ut{C}racker at {WNUT}-2020 task 2: Robustly
  identifying informative {COVID}-19 tweets using ensembling and adversarial
  training,'' in \emph{Proceedings of the Sixth Workshop on Noisy
  User-generated Text (W-NUT 2020)}.\hskip 1em plus 0.5em minus 0.4em\relax
  Online: Association for Computational Linguistics, Nov. 2020, pp. 404--408.
  [Online]. Available: \url{https://aclanthology.org/2020.wnut-1.57}
\BIBentrySTDinterwordspacing

\bibitem{pk-paw}
\BIBentryALTinterwordspacing
H.~Goyal, A.~Singh, and P.~Kumar, ``{PAW} at {S}em{E}val-2021 task 2:
  Multilingual and cross-lingual word-in-context disambiguation : Exploring
  cross lingual transfer, augmentations and adversarial training,'' in
  \emph{Proceedings of the 15th International Workshop on Semantic Evaluation
  (SemEval-2021)}.\hskip 1em plus 0.5em minus 0.4em\relax Online: Association
  for Computational Linguistics, Aug. 2021, pp. 743--747. [Online]. Available:
  \url{https://aclanthology.org/2021.semeval-1.98}
\BIBentrySTDinterwordspacing

\end{thebibliography}

\end{document}